\documentclass{article}
\usepackage[final]{nips_2016}

\usepackage[utf8]{inputenc} %
\usepackage[T1]{fontenc}    %
\usepackage{hyperref}       %
\usepackage{url}            %
\usepackage{booktabs}       %
\usepackage{amsfonts}       %
\usepackage{nicefrac}       %
\usepackage{microtype}      %
\usepackage{todonotes}
\usepackage{amsmath}
\usepackage{subcaption}

\usepackage{placeins}
\usepackage{tikz}
\usepackage{tikzscale}
\usetikzlibrary{arrows}

\usepackage{natbib}
\bibpunct{(}{)}{;}{a}{,}{,}

\newcommand{\M}{\mathcal{M}}
\newcommand{\St}{\mathcal{S}}
\newcommand{\A}{\mathcal{A}}
\newcommand{\regret}{\mathrm{Regret}}
\newcommand{\regretHAT}{\hat{\mathrm{Regret}}}

\title{Active Reinforcement Learning: \\ Observing Rewards at a Cost}

\author{
  David Krueger \\
  Montreal Institute for Learning Algorithms \\
  University of Montreal \\
  \texttt{david.krueger@umontreal.ca} \\
  \And
  Jan Leike \\
  Future of Humanity Institute \\
  University of Oxford \\
  \texttt{jan.leike@philosophy.ox.ac.uk} \\
  \And
  Owain Evans \\
  Future of Humanity Institute \\
  University of Oxford \\
  \texttt{owain.evans@philosophy.ox.ac.uk } \\
  \And
  John Salvatier \\
  AI Impacts \\
  \texttt{jsalvatier@gmail.com} \\
}

\begin{document}
\maketitle

\begin{abstract}
Active reinforcement learning (ARL) is a variant on reinforcement learning
where the agent does not observe the reward unless it chooses to pay
a \emph{query cost} $c > 0$.
The central question of ARL is
how to quantify the long-term value of reward information.
Even in multi-armed bandits, computing the value of this information
is intractable and we have to rely on heuristics.
We propose and evaluate several heuristic approaches for ARL
in multi-armed bandits and (tabular) Markov decision processes,
and discuss and illustrate some challenging aspects of the ARL problem.
\end{abstract}

%
\section{Introduction}

Translating human objectives into rewards for a reinforcement learning~(RL) agent is often challenging and time-consuming.
Tasks such as driving in a busy city or engaging in a conversation have complex reward structures which humans understand via high-level concepts such as ``safety'' and ``enjoyment''.
For such tasks, it may be prohibitively expensive to design a reward function by hand for an agent that does not already understand such concepts. %

Instead of specifying a reward function in advance, a human can evaluate state-actions as they occur and provide reward to the agent \emph{online}. %
In this case, it's not necessary to assign rewards to all possible state-actions, only to those which are visited.
Manually providing these rewards may still be labor intensive and costly, however.
For instance, in a clinical trial the rewards for an RL agent are patient outcomes, measured by expensive medical tests.
Likewise, it may be expensive to obtain high quality feedback from users of a web application.

To account for the cost of collecting reward feedback, we consider a simple modification to traditional RL called \emph{active reinforcement learning} (ARL).
In analogy with active learning, an ARL agent {\em chooses} when to observe the reward signal. 
\begin{quote}
 \textbf{Informal problem statement for ARL}: \newline
At each time-step, the agent chooses both an action and whether to observe the reward in the next time-step. 
If the agent chooses to observe the reward, then it pays the ``query cost'' c > 0.
The agent's objective is to maximize total reward minus total query cost.
\end{quote}

By intelligently choosing which rewards to observe, ARL agents can learn more efficiently about the reward function. 
As in active learning, the agent can exploit statistical dependencies between the rewards of different state-actions, observing rewards that are expected to be most informative about other rewards. 
Unlike in active learning, the goal in ARL is not to learn the reward function, but rather to take actions that maximize the discounted sum of rewards. %
Towards this end, ARL agents can also exploit knowledge of the environment's dynamics to query the state-actions most relevant to improving their policy.  

The ARL problem involves two important simplifying assumptions: 
\begin{enumerate}
\item Rewards are observed immediately (i.e. without any delay)
\item The agent only ever observes rewards for the {\em current} state-action (not past state-actions)
\end{enumerate}
Relaxing these assumptions would be valuable but is beyond the scope of this paper. %

For many problems, requiring a human to be ``on-call'', i.e.\ ready to provide rewards without delay, is itself costly.
This could be addressed by using a query cost which is situation dependent, e.g.\ determined online by a human.
However, when the agent acts on time-scales much shorter than the human, it's not possible to provide it instantaneous rewards. 
Further work could explore the consequences of reward signals being delayed.
Restricting queries to the current state-action is natural
if the agent's representation of state is missing some important features of the environment which the human can observe,
for example in partially observable problems.
In this setting, the human can make use of their additional information when providing rewards.
If the agent's state-representation captures all important features, then it could query the human about past or hypothetical state-actions. Future work will explore this case.

This paper presents preliminary work on active reinforcement learning~(ARL). 
We propose and evaluate several heuristic algorithms for ARL in multi-armed bandits and tabular MDPs. 
We present examples, theoretical observations and empirical comparisons that shed light on the character of the ARL problem. 
The central question of ARL is how to quantify the long-term value of reward information. 
As the cost of observing rewards gets high~(relative to possible discounted future rewards) the problem may become quite different from the traditional RL problem with fully-observed reward. 
\subsection{Related work}

The difficulty of encoding human objectives in reward functions has been articulated and addressed in a substantial literature. 
This includes work in inverse reinforcement learning \citep{ng2000algorithms, evans2015learning} and preference-based reinforcement learning~\citep{wirth2013preference}. 

The TAMER framework~\citep{knox2009interactively} has a similar motivation to ARL.
The authors consider the setting where rewards are provided online by a human ``teacher'' and explicitly take into account the time delays and noise in the human's responses. TAMER typically assumes that the human provides signals about the {\em value} of actions (e.g. $Q^*(s,a)$), rather than their rewards (as in ARL).

Recent work on ``Active Reward Learning''~\citep{DanielVMKP2014} also shares its motivation with ARL.
\citeauthor{DanielVMKP2014}\ test RL with online human feedback empirically, with humans evaluating how well a robot grasps objects.
Their algorithm is designed for continuous control tasks, where the reward function is defined on entire trajectories rather than individual actions.
They employ an active learning approach based on Gaussian Process regression and Bayesian Optimization.
However, their techniques are not readily applicable to the discrete problems we study here. 

Finally, there is a wide range of work in which an RL agent performs some kind of active learning of reward information in a setting that is less similar to ARL~\citep{weng2013interactive, lopes2014active, regan2011robust}.

\section{Active Bandits}

This section formally introduces and discusses
the active reinforcement learning problem
in the context of multi-armed bandits.

\subsection{Problem Formulation}

The \emph{active multi-armed bandit problem} is
an online learning problem
where the learner has to select from $K$ actions (`arms') in every round.
Choosing action $A_t \in \{ 1, \ldots, K \}$ in round $t$
yields a reward $R_t$ of
drawn from a time-independent distribution $\nu_{I_t}$
with mean $\mu_i$ unknown to the learner.
In contrast to the classical multi-armed bandit problem,
the learner does not observe the reward
unless they pay a fixed \emph{query cost $c > 0$}.
This cost is the same for every arm and every round.

Let $\mu^* := \max_i \mu_i$ denote the mean of the best arm.
The goal of the learner is to maximize expected reward up to a known horizon $n$,
or, equivalently, to minimize the \emph{expected regret}
\[
\regret_n = \mathbb{E} \left[ \sum_{t=1}^n (\mu^* - R_t + c Q_t) \right],
\]
where $Q_t$ is $1$ if the learner chooses to observe the reward in round $t$ and $0$ otherwise.
In other words, the expected regret is the expected reward
that is lost
because the learner did not blindly choose the best option in every round.

\subsection{Properties}

The active multi-armed bandit problem is
a subproblem of the partial monitoring problem~\citep{Piccolboni01}:
the action space consists of the $2K$ actions (the $K$ arms with and without querying) and
the feedback is the observed payoff or a blank symbol.
In contrast to most of the literature on partial monitoring,
we consider the environment to be stochastic instead of adversarial.
For distributions with finitely many possible rewards (such as Bernoulli distributions),
active multi-armed bandits are
stochastic finite partial monitoring problems~\citep{Komiyama15}.
In case of two outcomes,
(adversarial) partial monitoring problems can be classified in four different
categories: $0$, $\Theta(n^{1/2})$, $\Theta(n^{2/3})$, or $\Theta(n)$ regret~\citep{Antos13}.
These categories are called trivial, easy, hard, and hopeless respectively.
While multi-armed bandits fall into the easy category,
\emph{active} multi-armed bandits fall into the hard category:
their worse-case expected regret is
$\inf_\pi \sup_{\mu_1, \ldots, \mu_K} \regret_n \in \Theta(n^{2/3})$.

Intuitively, the fundamental difference to (regular) multi-armed bandit problems is in the cost of distinguishing two very close arms.
In active bandits, distinguishing two close arms incurs regret linear in the number of time steps needed to distinguish them ($n^{2/3}$ in the worst case), whereas in (regular) bandits the regret grows proportionally to the gap size ($n^{1/2}$ in the worst case).

To achieve optimal asymptotic regret, we can ignore the magnitude of the query cost, since it is constant.
Partial monitoring algorithms are designed for optimal asymptotic regret rate and do not take the magnitude of the query cost into account.
Here we are particularly interested in optimizing
the cost-dependent regret.

The optimal strategy is to query for a number of time steps
and then stop querying altogether.
If you don't query, then in the next step you face the same information,
except that the horizon is now shorter,
so the value of information can only have diminished.

Active multi-armed bandit problems degenerate into two familiar problems in the edge cases.
If $c = 0$ we face a regular bandit problem and the regret $\regret_n$ corresponds to the cumulative regret.
If $c \gg \max_i \Delta_i$ where $\Delta_i := \max_j \mu_j - \mu_i$ is the gap size,
then $\regret_n$ is dominated by the simple regret
(the gap of the chosen arm), and
so we have a best arm identification problem \footnote{For $c \gg 0$, it may be optimal to never query and simply chose the best arm according to the prior}.
But there is a trade-off between
simple regret and cumulative regret~\citep[Thm.~1]{Bubeck11}:
if the cumulative regret is small,
then there is a lower bound on the simple regret.
Any algorithm for a moderate cost setting therefore
has to manage the balancing act between cumulative and simple regret.

\subsection{Algorithms}
For finite sets of rewards,
\texttt{PM-DMED} achieves the optimal cost-independent regret rate~\citep{Komiyama15}.
We are looking for a (heuristic) algorithm that
\begin{itemize}
\item achieves optimal asymptotic (cost-dependent) regret rate,
\item performs well in practice, and
\item has constant or linear time complexity.
\end{itemize}
Ideally, the algorithm's performance does not degrade at the edge cases with $c = 0$ and $c \gg \max_i \Delta_i$.

A natural choice is an explore-then-exploit algorithm,
like for \emph{budgeted bandits}~\citep{Madani04}.
The central question is how to decide to stop querying.
In other words, what is the (long-term) value of information of an additional query?

If $\nu_1, \ldots \nu_K$ are Bernoulli distributions,
then the Bayes-optimal solution
can be found with dynamic programming:
The corresponding belief MDP has $\mathcal{O}(n^{2K})$ states
and can be solved in $\mathcal{O}(n^{2K})$ time steps.
This is polynomial for constant $K$,
but completely prohibitive in practice.

It is well-known that to solve partial monitoring problems
it is sometimes necessary to take actions that you strongly believe to be suboptimal to get information.
For active bandits, paying to see the reward is always suboptimal
by at least $c > 0$.
Nevertheless, if we never pay the cost, we learn nothing about the problem.
Because of this, strategies like optimism or Thompson sampling cannot tell us when to query. %
Moreover, there can be no index strategy%
\footnote{%
An index strategy computes for each arm
a number that is independent of the other arms
and takes the action with highest number.}
because the decision between two arms can depend
on the value of a third arm~\citep[Ex.~4]{Hay12}.
Lastly, active multi-armed bandits do not allow greedy solutions:
Usually one additional data point does not change our opinion
which arm we currently consider to be best.
A greedy strategy considers the value of information to be too low (or even zero) and stops querying prematurely~\citep[Sec.~5.2]{PowellRyzhov12}.
Therefore we have to take into account the long-range effects of querying for multiple time steps.

If $\nu_1, \ldots \nu_K$ are Gaussian distributions,
then we can estimate the value of paying the query cost for multiple time steps using
a knowledge gradient~\citep[Ch.~5]{PowellRyzhov12}.
As far as we know, how to efficiently compute the multi-step knowledge gradient for Bernoulli bandits is still an open question.

We introduce a new algorithm called \emph{mind-changing cost heuristic}~(\texttt{MCCH}).
Let $\hat{m}$ be an estimate of the expected number of time steps
we need to query in succession to move the posterior mean of the second best arm to the posterior mean of the best arm
(we use $\hat{m} := \max\{ 1, \lceil 2\min_i ((T_i + 1) \hat{\Delta}_i)^2 \rceil \}$).
Let $\regretHAT_n(i)$ denote the Bayes-expected regret
when committing to arm $i$ now (never querying again) and
let $\hat{i}$ denote the current estimate of the best arm.
The algorithm pays the query cost if and only if
\begin{equation}\label{eq:PRQ-criterion}
c\hat{m} < \alpha \regretHAT_n(\hat{i})
\end{equation}
where $\alpha > 0$ is a hyperparameter.
If \eqref{eq:PRQ-criterion} holds,
then the action is selected by some standard bandit algorithm,
such as \texttt{DMED}~\citep{Honda10}.
If \eqref{eq:PRQ-criterion} does not hold,
then the algorithm commits to the current best arm for the remaining time steps without paying the query cost.

\subsection{Experiments}

\begin{figure}[t]
\centering
\begin{subfigure}[b]{0.48\textwidth}
\centering
\includegraphics[width=\textwidth]{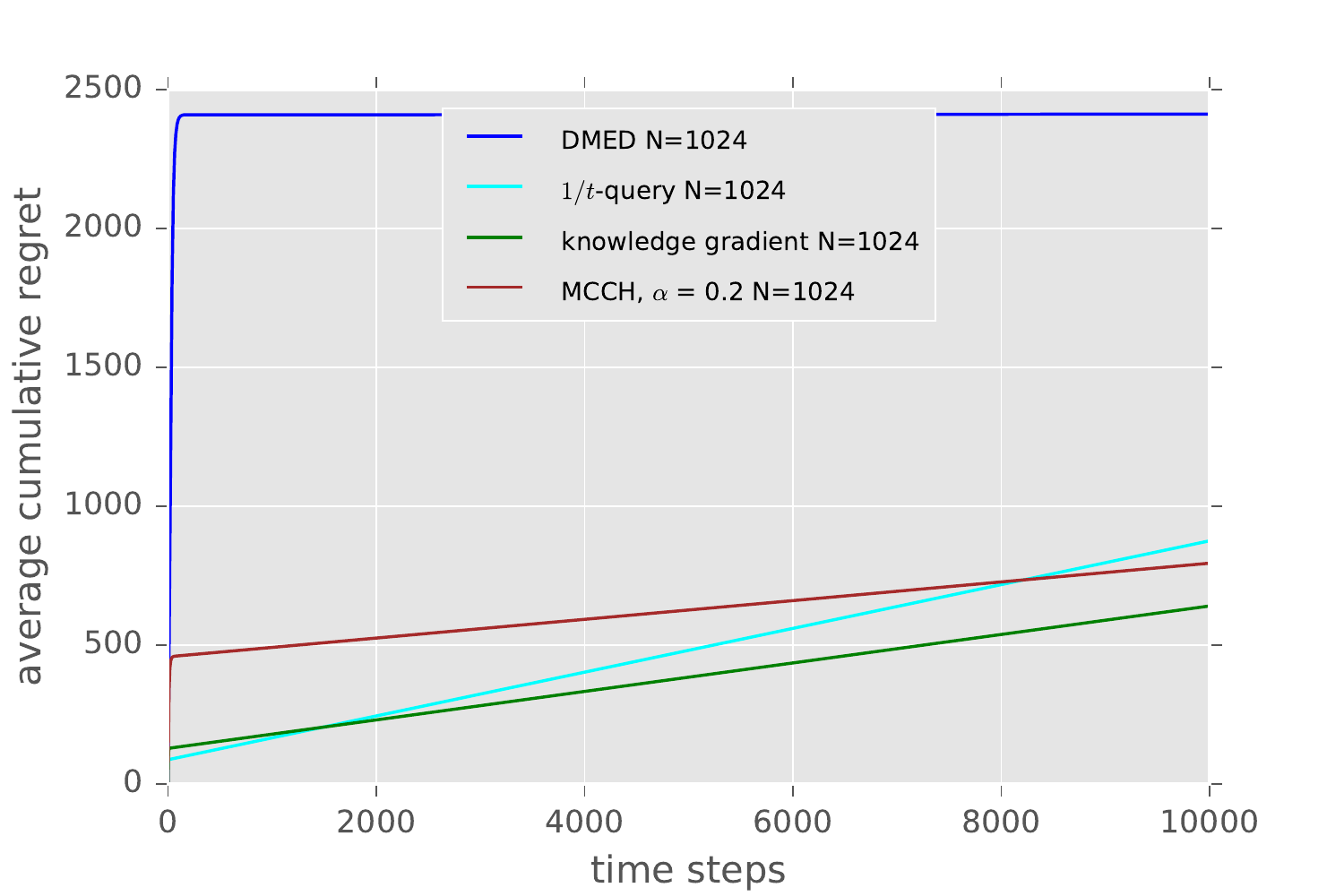}
\caption{Means 0.8 and 0.5, cost $c = 50$.}
\label{fig:regret50}
\end{subfigure}%
~ 
\begin{subfigure}[b]{0.48\textwidth}
\centering
\includegraphics[width=\textwidth]{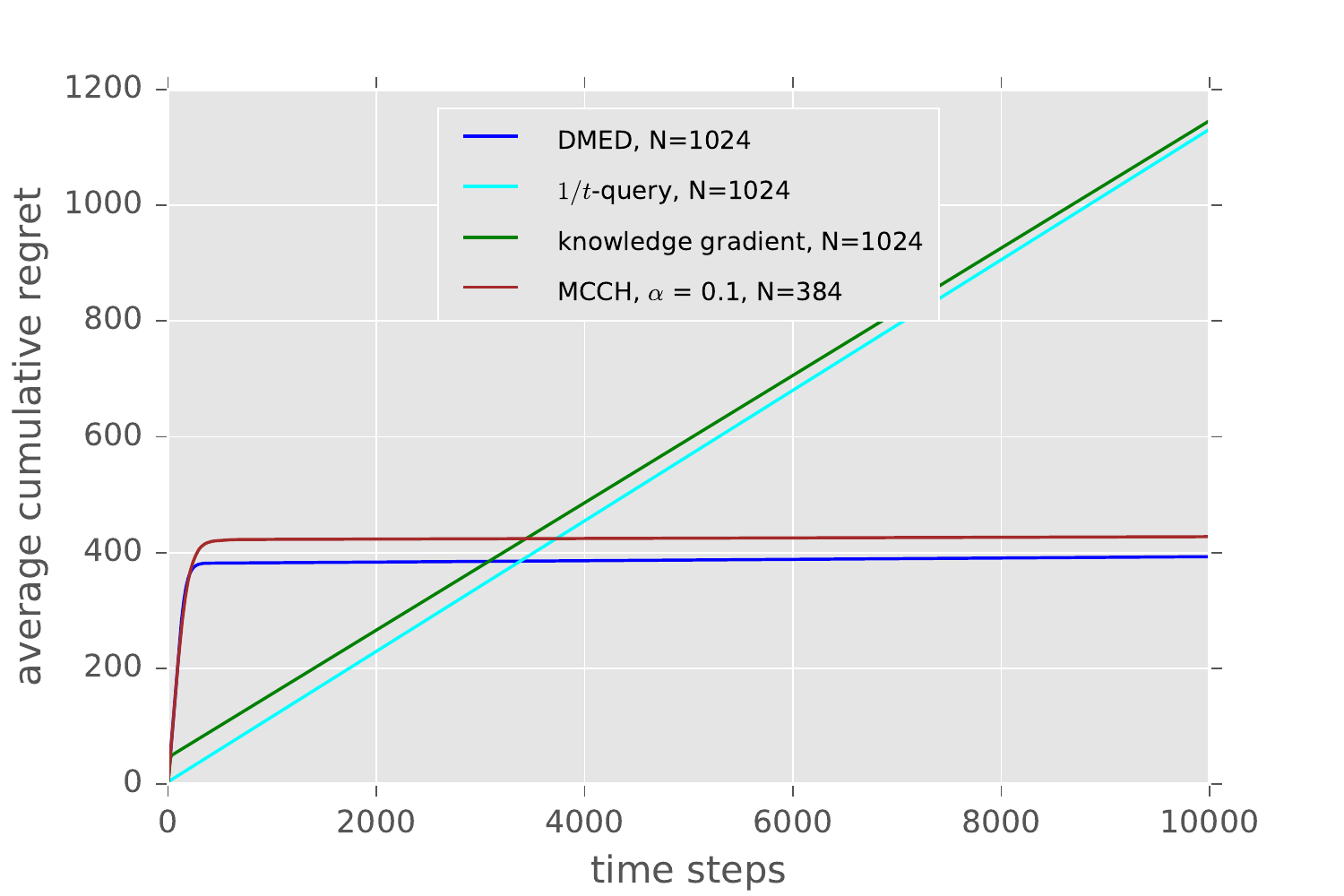}
\caption{Means $0.7, 0.5, 0.4, 0.4, 0.4, 0.4$, cost $c = 2$.}
\label{fig:regret2}
\end{subfigure}
\caption{
Average cumulative regret for different active bandit algorithms
for Bernoulli arms with horizon $10^4$.
We compare our algorithm \texttt{MCCH} with
knowledge gradient~\citep[Ch.~5]{PowellRyzhov12},
querying with probability $1/t$, and
a \texttt{DMED}~\citep{Honda10} variant that stops querying when
the algorithm selects only one arm.
The latter two algorithms do not take the query cost into account
and this is why they sometimes perform poorly.
}
\label{fig:regret}
\end{figure}

\begin{figure}[t]
\centering
\includegraphics[width=0.48\textwidth]{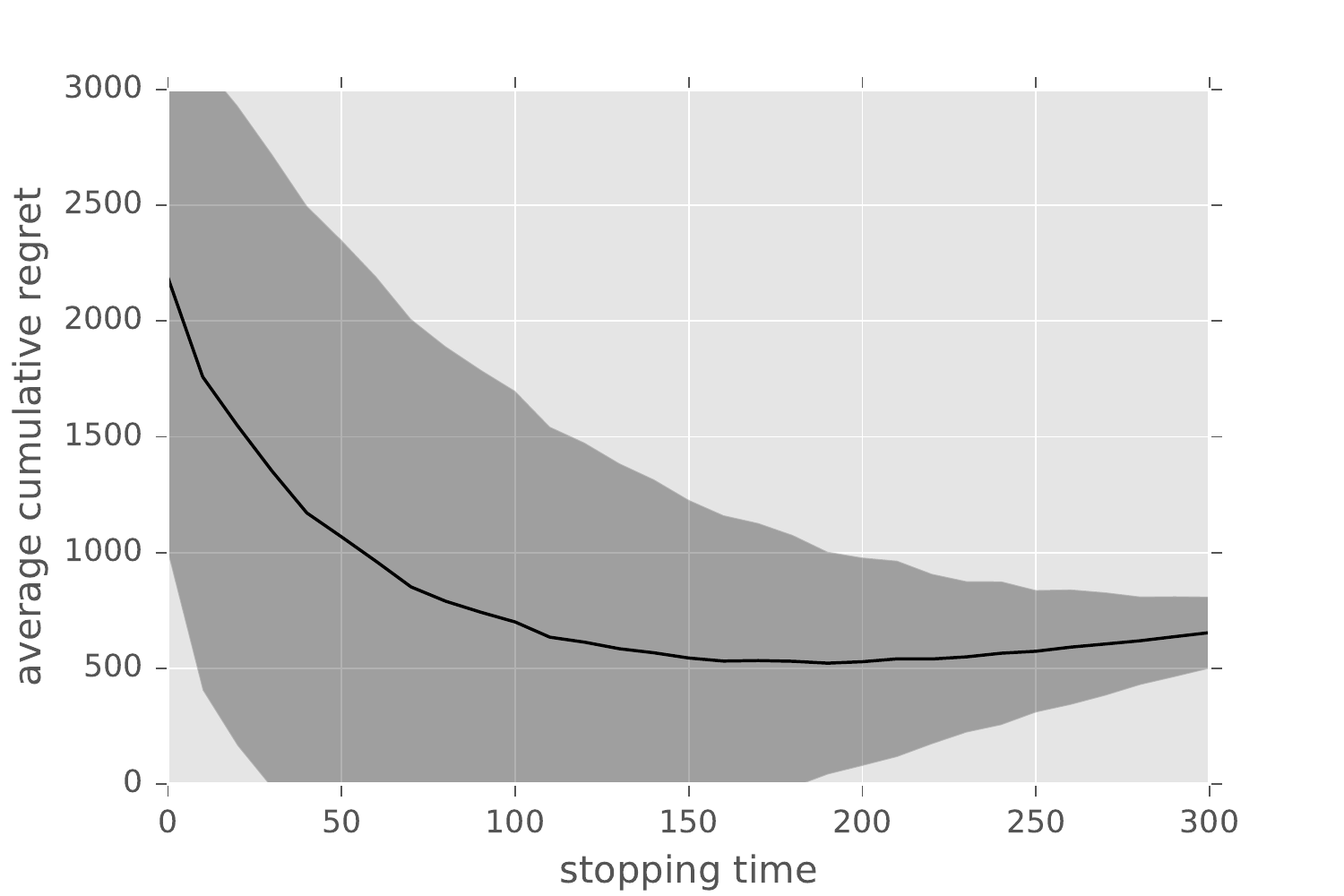}%
~
\includegraphics[width=0.48\textwidth]{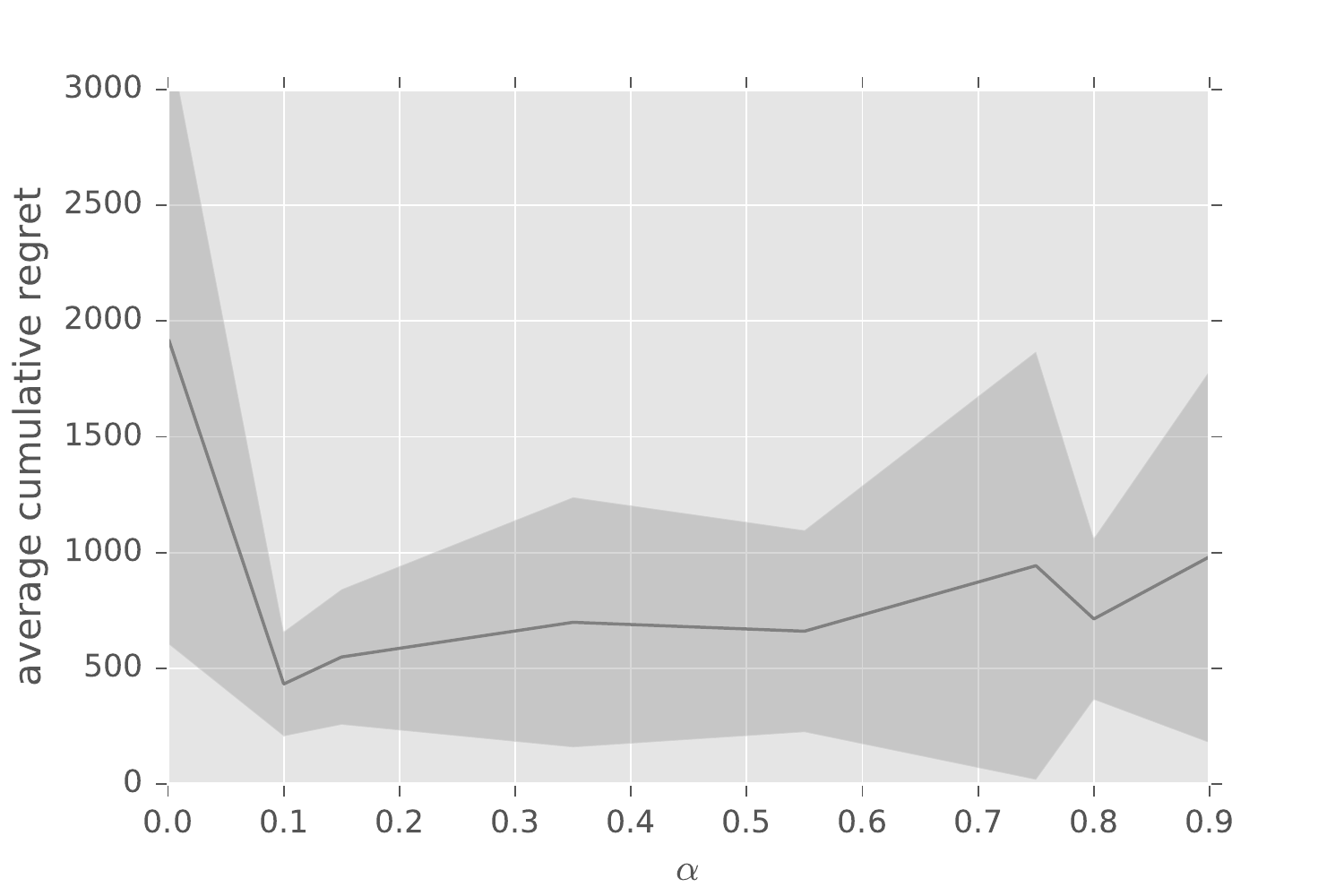}
\caption{Active 6-armed Bernoulli bandit with means
$0.7, 0.5, 0.4, 0.4, 0.4, 0.4$,
horizon $n = 10^4$, and cost $c = 2$.
We compare two policies:
\texttt{DMED} with a prespecified query stopping time (left)
and \texttt{MCCH}
for different values of the hyperparameter $\alpha$ (right).
The shaded area corresponds to one standard deviation.
\texttt{MCCH} achieves slightly better mean regret,
but also is much more robust to the choice of the hyperparameter.
}
\label{fig:alpha}
\end{figure}

\autoref{fig:regret} shows the average cumulative regret
of different active bandit algorithms for high ($c = 50$) and
moderate ($c = 2$) query costs.
All of these algorithms use \texttt{DMED}~\citep{Honda10}
to select actions.
To be able compute the non-greedy knowledge gradient policy,
we approximate prior, posterior, and likelihood with Gaussian distributions.
While \texttt{MCCH} never quite beats the other algorithms,
it performs more robustly for different query costs.
\autoref{fig:alpha} shows that the choice of its hyperparameter $\alpha$
is not very sensitive,
which makes it better suited than
an optimized fixed (problem-dependent) query stopping time.

\section{Active RL in MDPs}

This section studies Active Reinforcement Learning in tabular MDPs\footnote{
We use the notational standard MDPNv1 \citep{MDPNv1}
}. We begin by reviewing the problem. At every timestep $t$, the agent chooses an action $A_t \in \A$ and chooses whether to observe, or {\it query}, the reward sample $R_t$.
This decision is made after choosing the action but before seeing the next state.
Choosing to query incurs a cost, which we assume to be a constant $c>0$ that is known to the agent.
An ARL agent seeks to maximize the (discounted) sum of rewards (including unobserved rewards), minus the (discounted) sum of the query costs.

An optimal agent for ARL would query a state-action if the Expected Value of Sample Information (EVSI) \citep{EVSI} from the reward observation is larger than the query cost $c$. This is similar to the standard RL problem, where possibly sub-optimal actions are taken if their EVSI is high enough. However, ARL has some features that distinguish it from standard RL:

\begin{enumerate}
\item
In multi-armed bandits, the agent should only explore arms believed to be sub-optimal if they will query them. 
\item 
In MDPs, an agent with knowledge of the transition function can avoid querying certain state-actions all together. Some state-actions are \emph{unavoidable} on any policy. For example, the starting board in Chess is visited exactly once per game.

\item The agent can learn about the transition function $P$ without paying any query cost. This distinguishes model-learning from reward-learning. There are situations where the optimal agent learns about $P$ first (before learning anything about $R$) in order to direct subsequent queries more intelligently.
\end{enumerate}

\subsection{Algorithms}

We consider model-based algorithms for ARL with episodic, tabular MDPs. Our models are based on \emph{Posterior Sampling for Reinforcement Learning} ~\citep{Strens00}, an algorithm for episodic MDPs that is state-of-the-art in its empirical performance and its known regret bounds~\citep{osband2016posterior}. As with PSRL, our algorithms assume the agent has a Bayesian probability distribution on the reward function which is updated every episode. %
We follow PSRL in planning according to a reward function sampled from the posterior distribution (i.e.\ Thomson-sampling). The posterior distribution will also be used to estimate the value of querying. 

An obvious baseline is simply to use PSRL and query each state-action the first $N$ times it is visited. The question is how to choose the hyperparameter $N$. Our first algorithm provides one approach. 

\subsubsection{Simulating Query Strategies with Monte Carlo Rollouts}
The \emph{Simulated Query Rollout (SQR)} algorithm computes the average performance of a given query strategy~(e.g.\ a value for hyperparameter $N$) in environments sampled from the agent's posterior. This involves simulating the agent's performance across all the episodes for many different environments. This limits its use to choosing between a small set of possible query strategies. In our experiments, we use SQR to tune $N$, the number of times to query each state-action. We also consider an approximate version of SQR (\emph{ASQR}), which assumes the agent learns about queried rewards directly (rather than having to actually visit states to query them). By running ASQR at the start of every episode (\emph{ASQR in the loop}), the agent picks an $N$ tuned to the remaining time (rather than a fixed $N$ for the whole run).

\subsubsection{Estimating the Value of Information}

The algorithms above were based on the idea of picking an $N$ and querying each state-action $N$ times. Yet some state-actions are unavoidable (or hard-to-avoid on any policy) and these should never be queried.

Motivated by this observation about unavoidable states, we present algorithms based on approximating the Value of Information of each state-action $(s,a)$ independently. To estimate the value of knowing the reward $R(s,a)$, we compare the performance of ignorant vs. informed agents.  
The ignorant and informed agents may be, respectively:
\begin{enumerate}
\item The current agent, and the current agent with additional knowledge of $R(s,a)$ (``Greedy VOI'').
\item The agent which knows the value of $R$ everywhere except at $(s,a)$ (where it uses the current agent's beliefs), and the fully omniscient agent ("Omniscient VOI").
\end{enumerate}

We use the following greedy estimate of the value of information ($\mathit{VOI}$) which does not account for the possibility of further information gathering:
\begin{align}
\mathit{VOI} = \mathbb{E}_{P(\M)} [
    V_{\M} (\pi_{inf} )
  - V_{\M} (\pi_{ign} )
  ]
\end{align}
where $P(\M)$ is the agent's prior over environments, $V_{\M}(\pi)$ is the expected returns of policy $\pi$ in environment $\M$, and $\pi_{inf}, \pi_{ign}$ are the informed and ignorant agents' policies, respectively (in our case, the optimal policies in the informed / ignorant agents' expected environments).
Our algorithm computes $N(s,a)$ as a function of the $\mathit{VOI}$ of $R(s,a)$, estimated using Greedy VOI or Omniscient VOI, as well as 
the number of episodes remaining, $E$, the query cost, $c$, and a hyper-parameter, $k$, which controls the agent's eagerness to query:
\begin{align}
N(s,a) = \frac {k \cdot E \cdot \mathit{VOI}[R(s,a)] } {c} 
\end{align}

Note that we use the value of {\it perfect} information (EVPI), i.e.\ knowing $R(s,a)$, not of {\it sample} information (EVSI), i.e.\ observing some finite number of rewards sampled from $R(s,a)$, which we plan to explore next.
Besides using EVSI, these algorithms should also be extended to account for the opportunity cost of seeking to query a given state, which may be high if that state is hard to access, e.g.\ due to stochasticity in the environment.

\subsection{Experiments}

\begin{figure}[t]
\setlength{\belowcaptionskip}{-20pt}
\centering
\includegraphics[width=.75\textwidth]{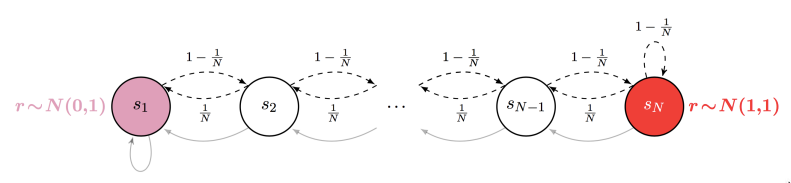}
\caption{The chain environment used in \citet{osband2016posterior}; our version has deterministic transitions.}
\label{fig:chain_env}
\end{figure}

We benchmark our proposed algorithms in the following 3 environments:
\begin{enumerate}
\item A chain of length 10, with reward only at the end  (see \autoref{fig:chain_env}).
\item A novel "long-Y" environment, also of length 10 (see \autoref{fig:y_env}). This environment has many unavoidable states not worth querying.
\item A 4x4 gridworld with rewards of $0$, $1/3$, $2/3$, $1$ along the diagonal and $0$ everywhere else.
\end{enumerate}

We compare for small ($c=1$) and large ($c=10$) query costs, and run for 4096 episodes \footnote{
Recall that the agent knows the number of episodes and uses this information to inform its query strategy.
}.
The agent has an independent standard normal prior for the reward of each state (or state-action, in the case of the chain) and learns only the mean (the variance is fixed to 1).
This prior is somewhat optimistic given the true reward structure.

We resample environments each episode for PSRL and to update our query strategy.
We use a single sampled environment for the Monte Carlo estimate of performance, although using more samples can significantly increase performance. %
For VOI algorithms, we set $k=1$. 
For the baseline algorithms, we either query each state-action up to 25 times (baseline 2), or query up to a total of $25 |\St| |\A|$ times (baseline 1); 
these numbers were tuned by hand based on previous experiments, which would not generally be possible in real applications.

Our results are presented in \autoref{fig:MDPs}.

As expected, the VOI algorithms perform better in the long-Y environment, as a result of not querying unavoidable states.
In the chain environment, however, the ASQR-based algorithms perform better; results in gridworld (not pictured) were similar.
Although the VOI algorithms have higher returns, they also query more, and hence perform worse overall; using smaller values of $k$ might improve their performance.
The VOI algorithms also query relatively less often on earlier episodes, which is undesirable.
This is because even when knowing $R(s,a)$ is valuable, in many sampled environments, it will not be.
Sampling more environments, or using an algorithm based on confidence-bounds would encourage earlier querying.
Not querying at all performs remarkably well in the chain environment when the query cost is high, likely due to the difficulty of the task and the optimism of the agent's reward prior.
Overall, these results demonstrate the value of updating query strategies online, and of differentially querying different state-actions, but should not be taken as a rigorous evaluation of the strength of the proposed algorithms.

%
%
%
%

%
%
%
%

%
\begin{figure}[t]
\centering
\resizebox{.55\textwidth}{!}{\begin{tikzpicture}
  [->,>=stealth',shorten >=1pt,auto,node distance=3cm,
  thick,
	main node/.style={circle,fill=black!0,draw,
  font=\sffamily\Large\bfseries,minimum size=15mm},
	goal node/.style={circle,fill=red!70,draw,
  font=\sffamily\Large\bfseries,minimum size=15mm}]

  \node[goal node] (R) {$r=1$};
  \node[main node] (L) [above of=R] {$r=0$};
  \node[main node] (s3) [left of=R] {$r=0$};
  \node[draw=none] (s2) [left of=s3] {$\ldots$};
  \node[main node] (s1) [left of=s2] {$r=0$};
  \node[main node] (s0) [left of=s1] {$r=0$};

  \path[every node/.style={font=\sffamily\small,
  		fill=white,inner sep=1pt}]
    (s0) edge [bend right=30] node[left=1mm] {$a_2$} (s1)
        edge [bend left=30] node[left=1mm] {$a_1$} (s1)
    (s1) edge [bend right=30] node[left=1mm] {$a_2$} (s2)
        edge [bend left=30] node[left=1mm] {$a_1$} (s2)
    (s2) edge [bend right=30] node[left=1mm] {$a_2$} (s3)
        edge [bend left=30] node[left=1mm] {$a_1$} (s3)
	(s3) edge [bend right=0] node[left=1mm] {$a_2$} (R)
        edge [bend left=0] node[left=1mm] {$a_1$} (L)
    (R) edge [loop right] node[left=1mm] {$a_1, a_2$} (R)
    (L) edge [loop right] node[left=1mm] {$a_1, a_2$} (L);
    \path (s1) -- node[auto=false]{\ldots} (s3);
\end{tikzpicture}}
\caption{The long-Y environment. Ideally, the agent should only query the two rightmost states, since the other states are unavoidable.}
\label{fig:y_env}
\end{figure}
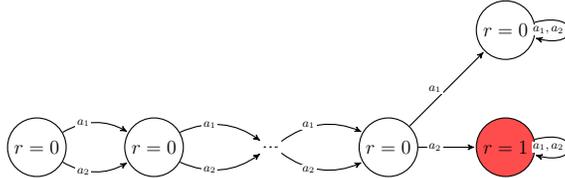
\begin{figure}[ht!]
\centering
\includegraphics[width=0.48\textwidth]{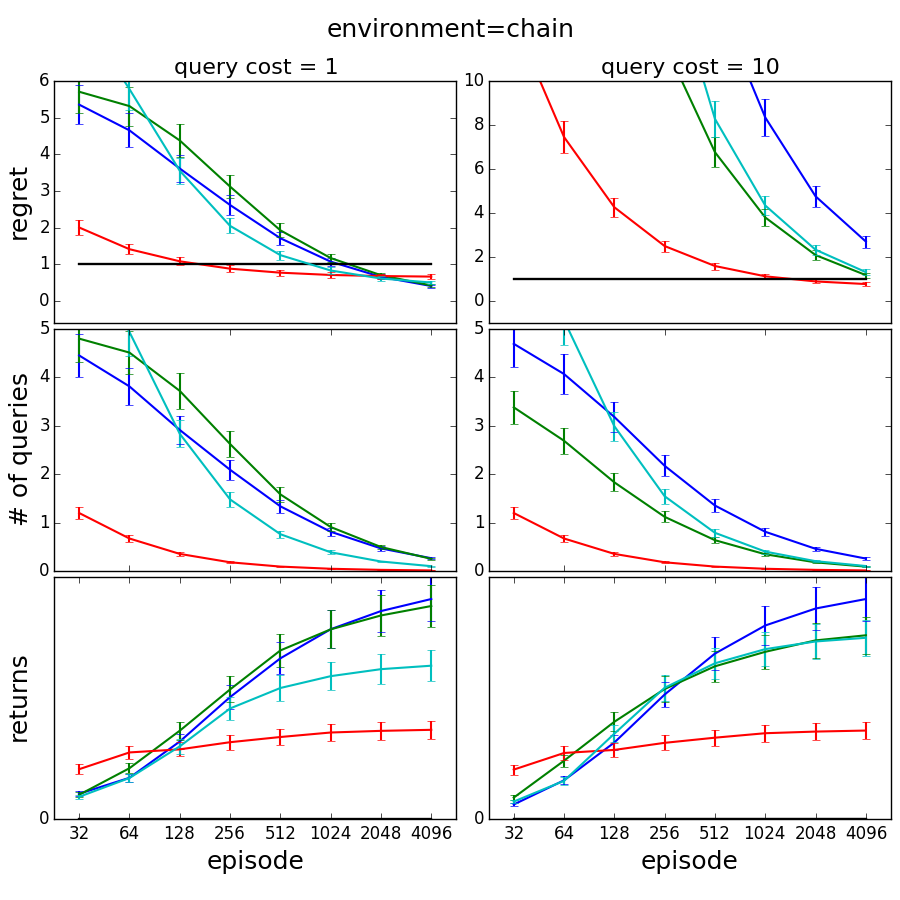}%
~
\includegraphics[width=0.48\textwidth]{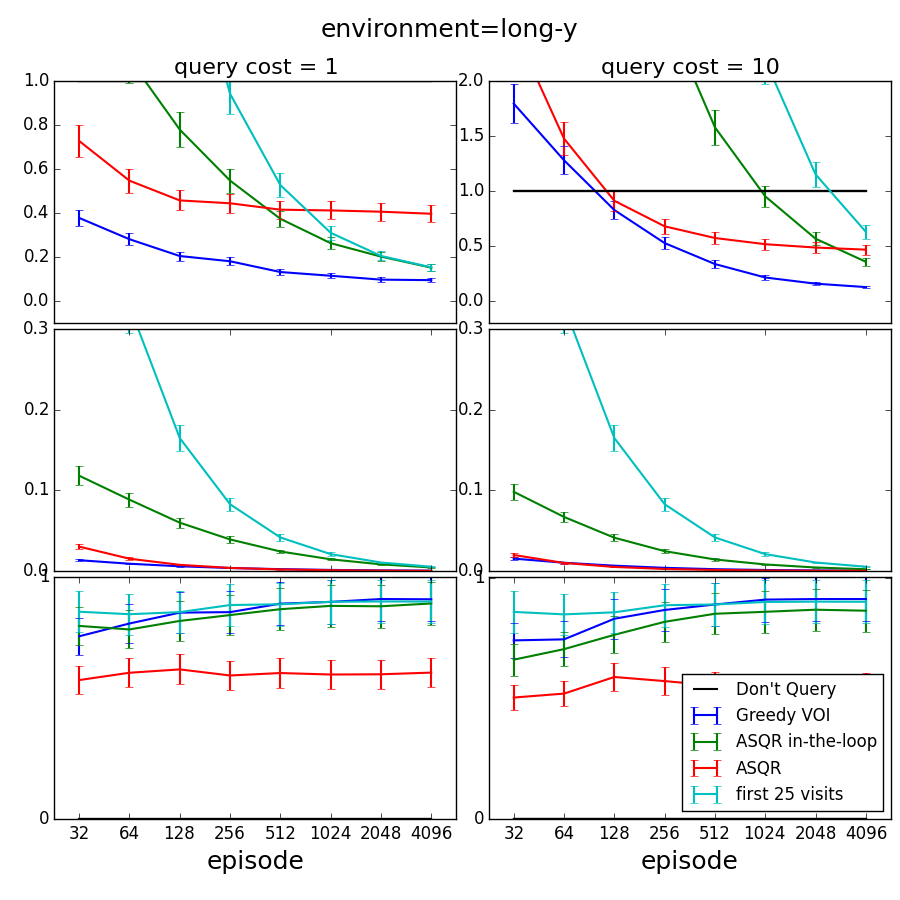}
\caption{
Average cumulative regret per episode (top), average number of queries per episode (middle), and average returns per episode (bottom) for different ARL algorithms in chain (left columns) and long-y (right columns) environments, with standard error bars.
}
\label{fig:MDPs}
\end{figure}

\FloatBarrier
\section{Conclusion}

We motivated and described Active Reinforcement Learning (ARL), in which an agent must pay to observe its reward signal. We demonstrated important properties that differentiate ARL from standard RL in both the multi-armed bandit and MDP cases. We explored a range of heuristic algorithms for these problems, some of which obtained promising empirical results (although our experiments were limited in scope).

Although we focused on the model-based case, where estimating performance of different query strategies is straightforward, we intend in future work to examine model-free approaches to ARL. 
Estimating the value of information or learning not to query unavoidable states seems challenging without a model, although visit counts could provide some relevant information.
Our current work focuses on multi-armed bandits and tabular MDPs with known dynamics.
Future work will investigate MDPs with unknown dynamics and large state spaces (for which function approximate is necessary). 
\paragraph{Acknowledgments.}
This work was supported by Future of Life Institute grant 2015-144846 (JS, DK, OE). 
We thank Andreas Stuhlm\"uller, Tor Lattimore, Reimar Leike, Ryan Lowe, Akram Erraqabi, and Jessica Taylor for helpful discussions.
The authors acknowledge the use of the University of Oxford Advanced Research Computing (ARC) facility in carrying out this work. %
\FloatBarrier
\bibliography{references}

\end{document}